\documentclass{article}
\usepackage[utf8]{inputenc}
\usepackage[T1]{fontenc}
\usepackage{spconf,amsmath,amssymb,graphicx,hyperref}

\usepackage{float}%提供float浮动环境
\usepackage{booktabs}%提供命令\toprule、\midrule、\bottomrule
\usepackage{multirow}%提供跨列命令\multicolumn{}{}{}

\title{Drag within Prior Distribution: \\Text-Conditioned Point-Based Image Editing within Distribution Constraints}
\name{HU HAOYANG$^{\dagger}$, Seo Masataka$^{\star}$, Yen-Wei CHEN$^{\dagger}$}
\address{$^{\dagger}$ Ritsumeikan University, Graduate School of Information and Engineering,\\
	$^{\star}$ Osaka Institute of Technology,}

\begin{document}
\ninept
\maketitle
\begin{abstract}
Diffusion-based point editing methods have gained significant traction in image editing tasks due to their ability to manipulate image semantics and fine details by applying localized perturbations on the manifold of noise latent. However, these approaches face several limitations. Traditional point-based editing relies on pairs of handle and target points to define motion trajectories, which can introduce ambiguity or unnecessary alterations. Furthermore, when the distance between the handle and target points is large, the accumulated perturbations often cause the noise latent deviation from inversion score trajectory, resulting in unnatural artifacts.
To address these issues in global editing tasks, we introduce a CLIP-based model to evaluate and guide intermediate editing steps, ensuring that the generated results remain both semantically aligned. Additionally, we propose a prior-preservation loss that constrains the optimized latent code to stay within the sampling space of the diffusion prior, improving consistency with the original data distribution, to ensure the model generates images along a familiar score trajectory. For fine-grained tasks, we present a directionally-weighted point tracking mechanism that steers the editing process toward the target direction within similar feature regions. This improves both the tracking accuracy and generation quality, while also reducing the editing time.

\end{abstract}
\section{Introduction}
\label{sec:intro}

The point-based method was first proposed in DragGAN\cite{draggan}, an approach that allows images to be freely edited by applying subtle adjustments to feature maps along the image manifold. This method enables zero-shot image editing within the allowable range of the manifold. However, due to the limitations of GAN model\cite{GANori,stylegan,stylegan2} size and generation quality, DragGAN has been replaced by point-based editing methods based on diffusion models\cite{ddpm,ddim,score} with higher generation quality\cite{stabledrag,shin2024instantdragimprovinginteractivitydragbased,choi2024dragtextrethinkingtextembedding,chen2024adaptivedragsemanticdrivendraggingdiffusionbased,clipdrag,fastdrag,freedrag,Xia2025DragLoRAOO,Lu2024RegionDragFR,Mou2024DiffEditorBA}, such as DragDiffusion\cite{dragdiffusion}, DragNoise\cite{dragnoise}, and GoodDrag\cite{gooddrag}.\\
In diffusion-based point editing methods, after performing DDIM inversion\cite{ddiminv} and latent optimization, the output image is sampled using a fine-tuned diffusion model with LoRA\cite{lora}. Although LoRA fine-tuning of the model is required for each image, which does not offer an advantage in generation time, the higher image generation quality still makes the diffusion-based point editing methods mainstream. Due to the robustness of the noise latent to minor perturbations, it usually generates high-quality images during simple semantic editing or short-distance editing.\\
In contrast to diffusion-based methods that adjust the global semantics of an image\cite{cfg}, diffusion-based point editing methods are more inclined to perform fine-grained modifications while preserving the overall semantic structure. However, in our experiments, we observe that point-based methods still possess a certain degree of global editing capability. Nevertheless, they face two major challenges:

\textbf{Prior distribution misalignment:} For a diffusion model to generate natural-looking images, the noise sample input must conform to the model’s expected prior distribution. While the inversion code obtained through DDIM inversion inherently satisfies this requirement, the edited inversion code—especially after long-range manipulations or substantial semantic shifts—tends to deviate from this prior. As a result, it no longer resembles a noisy latent sampled by encoding a natural image and injecting noise, which impairs the model’s generative fidelity.

\textbf{Local bias of U-Net and latent features:} Due to the inherent robustness of the U-Net’s intermediate features\cite{stylegan,stylegan2,draggan} and the noise latent to small perturbations in the latent space, diffusion-based point editing methods naturally bias the editing direction toward local fine-grained adjustments, regardless of whether the actual target is a global transformation or a semantic-level change. This local bias can lead to incorrect or undesired modifications, especially in complex editing tasks or when the user’s input is semantically ambiguous.\\
Therefore, the main contributions of our research are as follows:

We propose \textbf{DragPD}, a method that utilizes Prior-Preservation Regularization to keep the edited inversion code close to the prior. This enables our model to generate more natural and higher-quality images. At the same time, it also makes it possible to achieve longer-distance editing and more drastic semantic changes.

In the optimization step, we introduce a \textbf{CLIP-reward model}\cite{clip,xu2023imagerewardlearningevaluatinghuman,eyring2024renoenhancingonesteptexttoimage}. By evaluating the dynamic sampling results of the intermediate states, we obtained the reward score to guide the editing direction of the model in point-based editing, making the editing results point to the semantic expected by the user.

Many existing methods have introduced modifications to the motion supervision and point tracking processes. In our study, we observe that for large-scale semantic transformations, the proposed Prior Preservation Regularization (PPR), which constrains the optimization within the global feature space, yields significant improvements. On the other hand, for fine-grained local edits, we propose \textbf{Directionally-Weighted Point Tracking}, a method that adjusts the nearest neighbor search by incorporating angular similarity between candidate directions and the target point direction. This approach effectively reduces miss tracking\cite{freedrag} in subtle edits, where small spatial errors may lead to significant perceptual inconsistencies.

\section{methodology}
\label{sec:format}
\subsection{Point-based method}
Point-based image editing was initially introduced by DragGAN\cite{draggan}, which alternates iterative optimization within the latent space.

In point-based methods, the user give several pairs of editing points and target points, after performing DDIM inversion on the image, iterative optimization is carried out to gradually shift the image information around the editing points towards the target points.

During the optimization process, it consists of two alternating steps, motion supervision and point tracking.

\textbf{Motion Supervision}
Motion supervision specifies a handle point $p_i$ and a corresponding point $p_i + d_i$ that is moved one pixel toward the target point in the U-Net's intermediate feature map~\cite{draggan}. By computing the $\ell_1$ loss between two patches, the discriminative power of the U-Net’s intermediate features drives the latent space to update accordingly. In this way, the semantic information around the editing point gradually shifts toward the target location.

In diffusion-based point editing, we apply motion supervision to the intermediate layers of the denoising U-Net and backpropagate the resulting changes to the inversion code, enhancing the alignment between the U-Net features and the latent representation.

\begin{equation}
\begin{aligned}
\mathcal{L}_{\text{ms}}(z_t^k) =\ 
& \sum_{i=1}^n \sum_{q \in \Omega(p_i, r_1)}
\left\| F_{q + d_i}(z_t^k) - \text{sg}(F_q(z_t^0)) \right\|_1 \\
& + \left\| (z_t^k - \text{sg}(z_t^0)) \odot (\mathbb{1} - M) \right\|_1
\end{aligned}
\label{eq:ms_loss}
\end{equation}

Specifically, the first term aggregates the $\ell_{1}$ distance between the updated feature maps $F_{q}(z^{t}_{i})$ and the stop–gradient reference $sg(F_{q}(z^{0}_{i}))$ over all spatial locations, which encourages semantic consistency during editing. The second term acts as a complementary regularizer, suppressing deviations from the original latent $\hat{z}^{t}_{i}$ in regions not guided by motion, thereby stabilizing the optimization process. Together, these terms provide point–wise supervision that effectively propagates motion constraints through the diffusion trajectory.

\textbf{Point Tracking}
As motion supervision alters the latent content, the position of the edit point must be updated accordingly. Point tracking performs a nearest-neighbor search within a local region centered at the original edit point and selects the pixel whose feature value is closest to the original as the new edit point.

\subsection{Prior-Preserving Regularization}
In diffusion-based point editing methods, the input image is typically encoded into a noise map via DDIM inversion, ensuring that the resulting latent code \( z_{t}^{0} \) remains close to the prior distribution. However, after undergoing significant semantic modifications and global changes—as reflected in elevated CLIP scores—the updated latent code \( z_{t}^{k} \) often deviates from the prior distribution. This deviation disrupts the score-based sampling paths established by DDIM inversion and increases the risk of generating unnatural images.

Motion supervision can be interpreted as a mapping from the prior distribution to a post-editing distribution. The goal of \textbf{Prior-Preserving Regularization (PPR)} is to regulate this mapping using KL divergence, thereby encouraging the posterior distribution to remain close to the prior and enabling the diffusion model to generate natural-looking results.

However, a direct application of KL divergence typically requires explicit knowledge of the probability distributions involved—namely, the prior and posterior. In our case, the posterior is an \textit{empirical distribution} derived from the motion supervision mapping and is therefore not analytically tractable; its shape, mean, and variance cannot be directly computed.

Since our goal is not to force exact equivalence between the prior and posterior, but rather to maintain distributional proximity during iterative updates, a relaxed approximation suffices. While this approach is not theoretically exact, it is statistically meaningful and practically useful:

\textbf{Lemma 1.} The empirical distribution \( P(x_t \mid x_0) \) obtained by applying mild perturbations—such as motion supervision or reward loss over a small number of steps—to a normal distribution can still be approximated as a normal distribution due to the robustness of Gaussian noise.

\textbf{Lemma 2.} Let \( z_{t}^{0} \) and \( z_{t}^{k} \) be sample sets obtained via DDIM inversion and motion supervision, respectively. Then \( P(x_t) \)—the prior—is a known Gaussian, and by Lemma 1, \( P(x_t \mid x_0) \) can be approximated as Gaussian as well. Therefore, the means and variances of these two distributions can be estimated from \( z_{t}^{0} \) and \( z_{t}^{k} \), respectively.

Our objective is to regulate the motion supervision mapping such that the sample set \( z_{t}^{k} \), drawn from the empirical posterior \( P(x_t \mid x_0) \), does not deviate significantly from \( z_{t}^{0} \), which reflects the prior \( P(x_t) \). This helps ensure that the edited image remains within the natural image manifold. Importantly, we do not aim for complete distributional equivalence; overly strict regularization may hinder the semantic expressiveness of motion supervision. Despite the approximations in Lemmas 1 and 2, the proposed regularization strategy is both principled and effective.
Therefore, prior preservation normalization is ultimately implemented via the KL divergence between two small-sample distributions, $z_{0t}$ and $z_{kt}$:
\begin{equation}
\mathcal{L}_{\mathrm{KL}} = \mathrm{KL}\left( \mathcal{N}(\mu_t^k, \sigma_t^k) \,\|\, \mathcal{N}(\mu_t^0, \sigma_t^0) \right)
\label{eq:kl_loss}
\end{equation}
where $\mu_t^0$ and $\sigma_t^0$ denote the mean and standard deviation of the inversion latent $z_t^0$, and $\mu_t^k$ and $\sigma_t^k$ represent the mean and standard deviation of the optimized latent $z_t^k$ after $k$ steps of editing.
\begin{figure}[t]
\centering
\includegraphics[width=0.9\columnwidth]{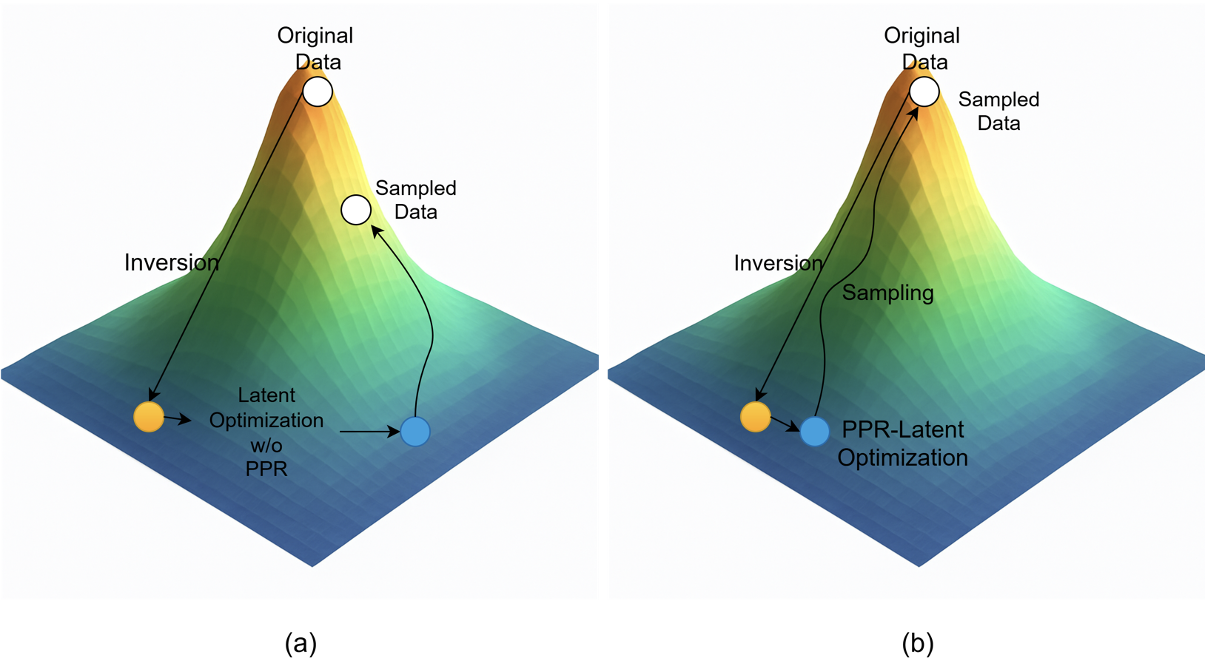} % Reduce the figure size so that it is slightly narrower than the column. Don't use precise values for figure width.This setup will avoid overfull boxes.
\caption{In diffusion-based point editing methods, image reconstruction relies on a \textbf{reversible generative trajectory} modeled from a single input image. During long-range manipulations or those guided by strong text prompts, the edited noise latent often deviates significantly from the prior distribution obtained via inversion. Our PPR explicitly regularizes the latent space during editing, encouraging the manipulated latent to remain close to the original prior distribution. }
\label{fig1}
\end{figure}

\subsection{Enhanced Text Guidance}
Motion Supervision has become a popular technique in image editing due to its high degree of editing freedom. However, it still faces two difficult limitations:

1. Sometimes, it is difficult to achieve the expected editing results merely based on the handle point and target point. Motion supervision often struggles to distinguish between local and global targets. 

2. Due to the robustness of the intermediate layer features of U-net and the noise latent to the image manifold, when dealing with unclear targets and multiple semantic interpretations, the gradient of motion supervision will naturally decline in the direction close to the image manifold, even if this is not the editing direction expected by the user.

Therefore, we consider introducing text guidance based on reward feedback to control the gradient direction of motion supervision. We use the results of the clip and contrastive loss function as the reward score. The mathematical formula of the contrastive loss function is as follows:\\
\begin{equation}
\mathcal{L}_{\text{clip}} = 1 - 
\frac{\langle E_I F(z_t^k), E_T(P_t) \rangle}{\| E_I F(z_t^k) \cdot E_T(P_t) \|} 
+ \lambda\frac{\langle E_I F(z_t^k), E_T(P_I) \rangle}{\| E_I F(z_t^k) \cdot E_T(P_I) \|}
\label{eq:kl_loss}
\end{equation}

Let $F$ denote the denoising UNet combined with a VAE decoder, responsible for reconstructing the image from the latent space. $E_I$ represents the CLIP image encoder, which embeds visual features into a shared semantic space. $z_t^k$ denotes the intermediate latent code obtained at the $k$-th optimization step during the inversion process. $E_T$ refers to the CLIP text encoder, which embeds textual prompts. $P_t$ and $P_I$ represent the prompts corresponding to the target and the initial semantic concepts, respectively. Both are derived from user-provided inputs.

\subsection{Directionally-Weighted Point Tracking}
\begin{figure}[h]
\centering
\includegraphics[width=0.9\columnwidth]{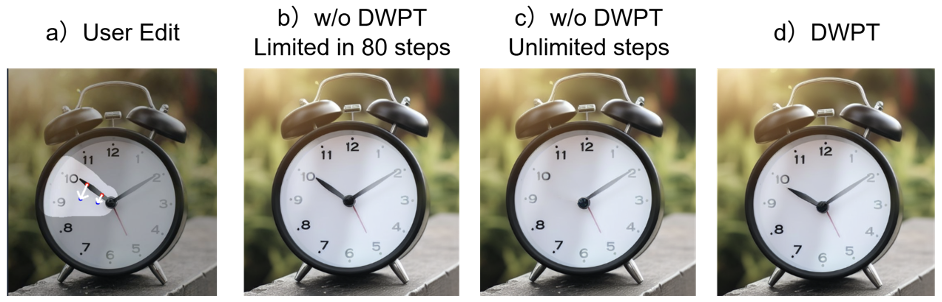} % Reduce the figure size so that it is slightly narrower than the column. Don't use precise values for figure width.This setup will avoid overfull boxes.
\caption{Common failure cases in fine-grained point-based editing.(b): Handle point detaches from local features, leading to ineffective editing.
(c): Detached handle point oscillates under repeated motion supervision, causing local overfitting.
(d): Our Directionally-Weighted Point Tracking (DWPT) keeps the handle aligned with the target direction, preventing feature detachment and overfitting.}
\label{fig2}
\end{figure}
In original point tracking, the nearest neighbor is selected based solely on feature distance. However, this may lead to incorrect matches in ambiguous regions or when multiple similar features patches exist. Existing approaches have proposed improvements to point tracking, such as restricting the search space to a local region\cite{clipdrag} or advancing along a predefined directional vector\cite{freedrag}. However, these strategies often constrain the flexibility of point tracking, making it less responsive to global gradient variations. As a result, the tracker may fail to accurately follow meaningful feature shifts across the entire image, especially in cases where large-scale semantic transformations are involved.

To address this, we incorporate directional information by prioritizing candidate points whose displacement vectors are more aligned with the intended motion direction toward the target point.

Given the current handle point $\mathbf{p}_i$ and the corresponding target point $\mathbf{t}_i$, we first define the direction vector as:
\begin{equation}
\mathbf{d}_i = \frac{\mathbf{t}_i - \mathbf{p}_i}{|\mathbf{t}_i - \mathbf{p}_i| + \epsilon}
\end{equation}
where $\epsilon$ is a small constant to avoid division by zero. For each candidate point $\mathbf{q}_j$ in the neighborhood of $\mathbf{p}_i$, we compute its direction vector $\Delta \mathbf{q}_j = \mathbf{q}_j - \mathbf{p}_i$, normalize it, and calculate the cosine similarity to $\mathbf{d}_i$:
\begin{equation}
\cos \theta_j = \frac{\Delta \mathbf{q}_j \cdot \mathbf{d}_i}{|\Delta \mathbf{q}_j| + \epsilon}
\end{equation}
We then define an angular weighting factor:
\begin{equation}
w_j = \lambda \cdot \cos(\theta_j) + (1 - \lambda)
\end{equation}
The candidate feature distance \(\mathcal{D}_j\) is adjusted using a directionally-aware weight:
\begin{equation}
\tilde{\mathcal{D}}_j = \frac{\mathcal{D}_j}{w_j}
\end{equation}

Here, \(\mathcal{D}_j\) denotes the L1 distance between feature embeddings, and \(\theta_j\) is the angle between the candidate vector and the desired motion direction. The hyperparameter \(\lambda \in [0, 1]\) controls the relative importance of directional alignment.

Unlike hard-thresholding strategies that entirely discard candidates based on angular deviation, this soft weighting scheme provides smoother optimization dynamics and greater robustness to local feature noise. Candidates that deviate slightly from the target direction are still considered, albeit with reduced weight, which helps avoid sudden tracking failures and ensures stable semantic progression.

In our experiments, we set \(\lambda = 0.05\), which balances directional bias and robustness.

\section{experiments}
\subsection{Compare with SOTA methods}
\begin{figure}[h]
\centering
\includegraphics[width=0.9\columnwidth]{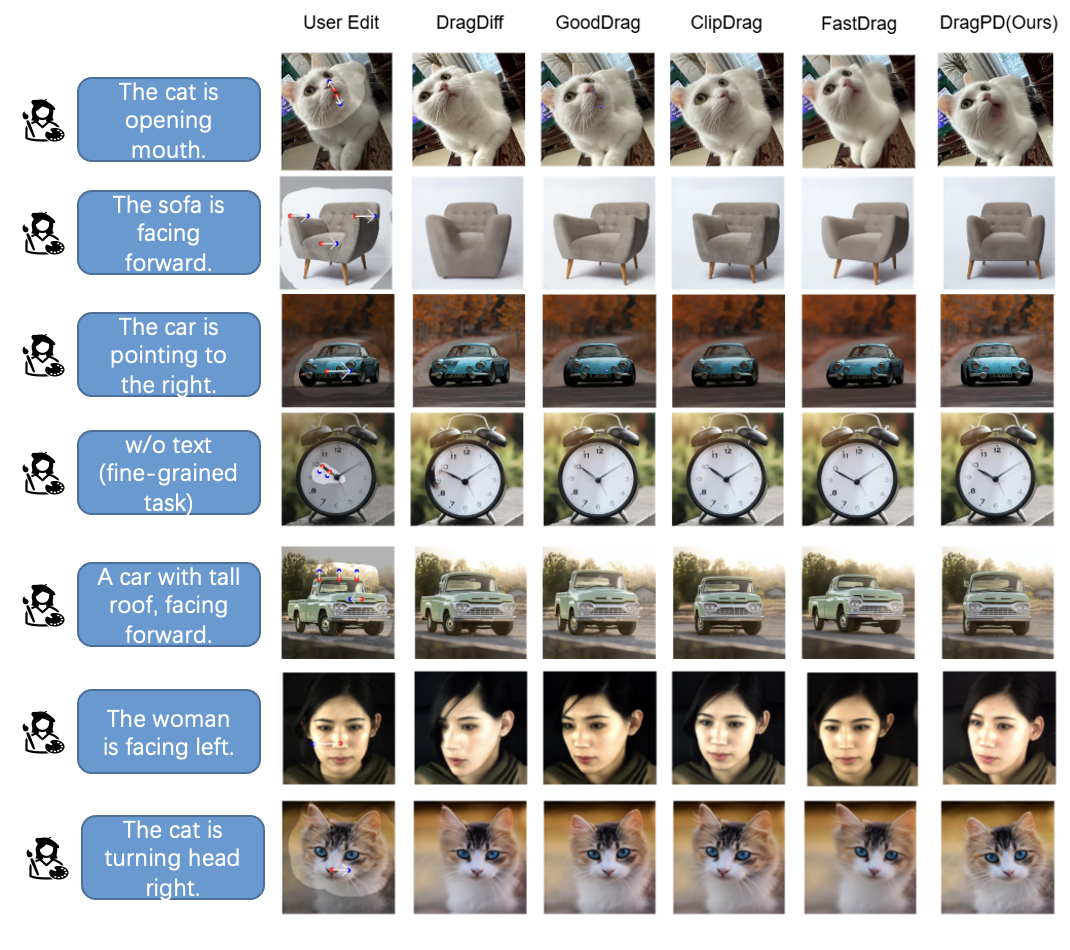} 
\caption{We present a subset of comparative results with diffusion-based drag editing methods, evaluated on both our custom dataset and the Drag100 benchmark. }
\label{fig3}
\end{figure}
To validate the effectiveness of our proposed method, we design a set of comparative experiments focusing on Prior-Preserving Regularization (PPR). Specifically, we compare our approach with several state-of-the-art diffusion-based point editing methods, including DragDiffusion, GoodDrag, FastDrag, StableDrag, and SDE-Drag.

We also employ multiple evaluation metrics that measure both image fidelity and alignment with target semantics. These comparisons allow us to quantitatively and qualitatively demonstrate the advantages of our proposed DragPD framework.

In our experiments, we adopt \textit{Stable Diffusion 1.5}\cite{ldm} as the generative backbone, which is fine-tuned using a Rank-16 LoRA configuration. During the optimization stage, we adopt CLIP-ViT-B/16\cite{vit} as reward model, set the weight of the CLIP reward to \( \lambda_{\text{CLIP}} = 150 \), and the weight of the Prior-Preserving Regularization (PPR) to \( \log(\lambda_{\text{KL}}) = 4 \). These hyperparameters are chosen to achieve a balance between image fidelity and the extent of semantic transformation toward the target prompt.

To evaluate the performance of our method, we adopt CLIP scores and LPIPS\cite{lpips} as complementary metrics. CLIP scores\cite{topiq} with initial content prompt $\mathcal{CLIP}_{object}$ and target prompts $\mathcal{CLIP}_{target}$ are used to assess the extent of semantic change and the ability to preserve original content. To evaluate the efficiency of latent optimization, we adopt \textit{mean distance} which calculates the average Euclidean distance between the handle points and their corresponding target points within a limited 80 drag steps.
\label{sec:pagestyle}
\begin{table}[h]
\centering
\caption{Quantitative comparison on the mix of DragBench and our data between SOTA dragging-based image editing methods.}
\begin{tabular}{ccccc} % 指定表格列数和对齐方式
\toprule % 顶部横线
\midrule
Model & 1 - LPIPS$\uparrow$ & MD$\downarrow$ & $\mathcal{CLIP}_{obj}\uparrow$ & $\mathcal{CLIP}_{tar}\uparrow$ \\ % 表头
\midrule % 中间横线
DragDiff      & 0.803 & 41.42  & 0.277 & 0.282   \\
StableDrag     & 0.877 & 40.05 & 0.281 & 0.290   \\
GoodDrag           & 0.894 & 29.96 & 0.280 & 0.296    \\
FreeDrag           & 0.822 & 39.77 & 0.275 & 0.283  \\
AdaptiveDrag       & 0.862 & 36.25 & 0.277 & 0.280   \\
CLIPDrag           & 0.880 & 45.87 & 0.282 & 0.304 \\
SDE-Drag       & 0.899 & 42.40 & 0.273    & 0.280   \\
FastDrag           & 0.787 & 60.86 & 0.285    & 0.278      \\
\midrule
\textbf{Ours w/o text}  & \textbf{0.901} & \textbf{20.23} & \textbf{0.291} & 0.301  \\
\textbf{Ours}      & 0.892 & 22.26 & 0.288 & \textbf{0.316}  \\
\midrule % 底部横线
\bottomrule
\label{Table1}
\end{tabular}
\end{table}

As shown in Table \ref{Table1}, the DragPD variant without the reward model already achieves strong performance in both image fidelity and alignment with the original semantics, benefiting from the use of Prior-Preserving Regularization (PPR) and Directionally-Weighted Point Tracking (DWPT). The introduction of DWPT notably reduces the mean distance (MD) between handle and target points. When the reward model is incorporated, although the image fidelity slightly decreases—mainly due to the influence of the global loss on the entire image and mask regions—the alignment with the target semantics improves significantly. This enhancement enables DragPD to accomplish challenging edits that are difficult for traditional point-based editing methods.

\begin{table}[h]
\centering
\caption{Optimization time comparison with outperforming SOTA methods}
\begin{tabular}{ccccc} % 指定表格列数和对齐方式
\toprule % 顶部横线
 & DragDiff & GoodDrag & CLIPDrag & \textbf{Ours w/o text} \\ % 表头
\midrule % 中间横线
time $\downarrow$      & 49.8s & 66.3s  & 151.0s & \textbf{43.1s}   \\
\bottomrule % 底部横线
\label{Table2}
\end{tabular}
\end{table}

As shown in Figure~\ref{fig3}, our method demonstrates significant advantages in both global semantic change tasks and fine-grained local editing scenarios. By jointly leveraging prior-preserving regularization and directionally-guided point tracking, our approach achieves more accurate semantic alignment and faster convergence across a wide range of editing complexities.

Table \ref{Table2} presents a comparison of the optimization time among several top-performing SOTA methods in the quantitative evaluation. It can be concluded from the table that when Text guidance is not used, our method achieves a certain improvement in generation time while maintaining the generation quality. This is mainly attributed to DWPT solving the problem of repeated drift of handle points and pruning the optimization steps to a certain extent.

\subsection{Ablation Results}
\label{sec:typestyle}

\textbf{w/o PPR}
Since the Diffusion-based Point Editing method relies on LoRA to restore image information, the image restoration process follows a similar score path. Applying CLIP-Reward for image editing without PPR (Prior Preservation Regularization) results in significant changes to the entire image, causing the image distribution to deviate significantly from the prior, and also leading to a deviation in the score path during the restoration process. As shown in the Figure~\ref{fig3} (c), the model generates unnatural images without the use of PPR.\\
\textbf{w/o CLIP-Reward}
Applying PPR alone can produce natural results with semantic variations. However, in complex editing scenarios that may partially exceed the tolerance of the image manifold such as changing a cat’s mouth from closed to open—the absence of CLIP-reward gradients makes it difficult to achieve the intended manipulation.

\begin{figure}[h]
\centering
\includegraphics[width=0.9\columnwidth]{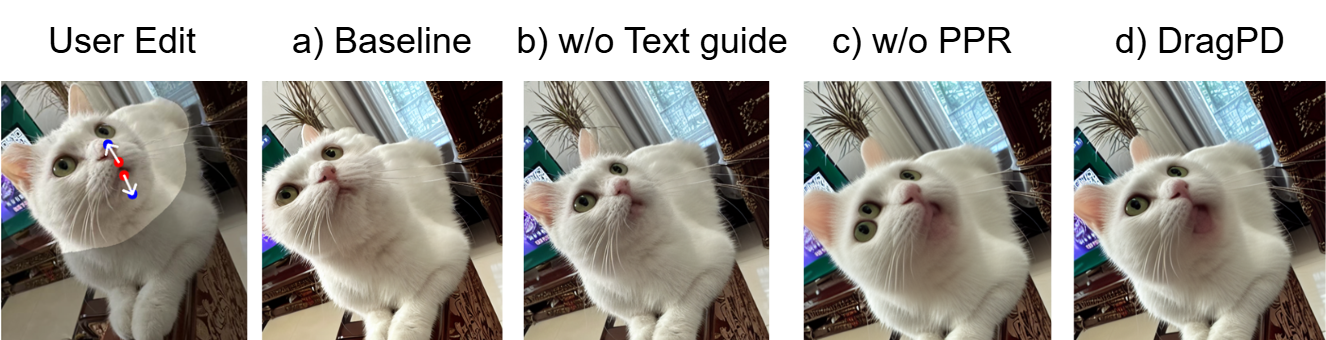} 
\caption{We conduct ablation studies on representative tasks characterized by ambiguous editing goals and global semantic shifts that approach the boundary of the image manifold. }
\label{fig4}
\end{figure}

\section{Conclusion}
\label{sec:copyright}
We propose Prior-Preservation Regularization (PPR) and CLIP-based reward from the perspectives of global and local optimization, respectively—PPR constrains prior deviation in the optimization mapping, while CLIP-reward enables more extensive global transformations. Additionally, we enhance the precision of local editing through Directionally-Weighted Point Tracking (DWPT). By integrating these components, our DragPD model achieves \textbf{a well-balanced trade-off among controllability, consistency, and robustness in image editing tasks}. 

\newpage

\bibliographystyle{IEEEbib}
\bibliography{strings,refs}

\end{document}